\RequirePackage{fix-cm}
\documentclass[twocolumn]{svjour3}          
\smartqed  
\usepackage{graphicx}
\usepackage{nicefrac}
\usepackage{makecell}

\usepackage[caption=false]{subfig}

\usepackage{threeparttable}
\usepackage{subfig}

\usepackage{latexsym}
\usepackage{amsmath}

\journalname{Journal of the Brazilian Society of Mechanical Sciences and Engineering}
\begin{document}

\title{Estimation of Tire-Road Friction for Road Vehicles: a Time Delay Neural Network Approach}
	



%

\author{Alexandre M. Ribeiro \and Alexandra Moutinho \and Andr\'e R. Fioravanti \and Ely C. de Paiva}

\authorrunning{Alexandre M. Ribeiro \and Alexandra B. Moutinho \and Andr\'e R. Fioravanti \and Ely C. de Paiva}

\institute{Alexandre M. Ribeiro \at
			Department of Computational Mechanics,  School of Mechanical Engineering, Mendeleyev Street 200, 13083-860 Campinas, Brazil \\
			\email{amribeiro@fem.unicamp.br}
			\and
			Alexandra Moutinho \at
			Institute of Mechanical Engineering - IDMEC, LAETA, Instituto Superior T\'ecnico, Universidade de Lisboa, Av. Rovisco Pais, 1049-001 Lisbon, Portugal\\
			\email{alexandra.moutinho@tecnico.ulisboa.pt}
			\and
			Andr\'e R. Fioravanti \at
			Department of Computational Mechanics,  School of Mechanical Engineering, Mendeleyev Street 200, 13083-860 Campinas, Brazil \\
			\email{fioravanti@fem.unicamp.br}
			\and
			Ely C. de Paiva \at
			Department of Integrated Systems,  School of Mechanical Engineering, Mendeleyev Street 200, 13083-860 Campinas, Brazil \\
			\email{elypaiva@fem.unicamp.br}
}

\date{Received: date / Accepted: date}

\maketitle

\begin{abstract}
The performance of vehicle active safety systems is dependent on the friction force arising from the contact of tires and the road surface. Therefore, an adequate knowledge of the tire-road friction coefficient is of great importance to achieve a good performance of different vehicle control systems. This paper deals with the tire-road friction coefficient estimation problem through the  knowledge of lateral tire force. A time delay neural network (TDNN) is adopted for the proposed estimation design. The TDNN aims at detecting road friction coefficient under lateral force excitations avoiding the use of standard mathematical tire models, which may provide a more efficient method with robust results. Moreover, the approach is able to estimate the road friction at each wheel independently, instead of using lumped axle models simplifications. Simulations based on a realistic vehicle model are carried out on different road surfaces and driving maneuvers to verify the effectiveness of the proposed estimation method. The results are compared with a classical approach, a model-based method modeled as a nonlinear regression.
\keywords{Road friction estimation \and Artificial neural networks \and Recursive least squares \and Vehicle safety \and Road vehicles}
\end{abstract}

\section{Introduction}
\label{intro}
One of the primary challenges of vehicle control is that the source of force generation is strongly limited by the available friction between the tire tread elements and the road. In order to better understand vehicle handling due to force generation mechanisms, several research activities related to vehicle dynamics and control fields are oriented towards estimation of all components of the tire-ground contact.

The knowledge of specific tire-road contact operation points, such as the saturation point where the tire generates the maximum force available from friction, could lead to a new range of applications in vehicle control. Moreover, current commercial vehicle safety systems, such as Anti-lock Brake Systems (ABS), Traction Control Systems (TCS) and Electronic Stability Control (ESC) could have a significant improvement in performance by the knowledge of the full vehicle states and operating conditions that are still limited by the lack of information \cite{singh2014Estimation}. Therefore, to reach this full potential the recognition of the tire's limit handling is indispensable. With this is mind, we emphasize the importance of estimating the vehicle road conditions, specifically the Tire Road Friction Coefficient (TRFC).

Friction estimation often relies on a model-based estimator using a well-defined and interpretable mathematical model for the purpose of capturing the inherent friction effects under the tire dynamics (forces and moments). The most common model-based approaches use the steering system model \cite{Liang2019,hsu2006method}, quarter-car model \cite{AlZughaibi2018}, four-wheel vehicle dynamic model \cite{ahn2011robust}, powertrain and wheel dynamics model \cite{Castro2010}.

For the estimation problem, in \cite{pasterkamp1997} the correlation between the self-aligning moment of the steering wheels and road friction was firstly examined. Lately, \cite{hsu2009estimation,hsuJournal,ahn2011robust,ahnJournal2013} explored the use of a nonlinear Recursive Least Squares method employed as a mean for identification of tire-road friction through an observed data composed by the self-aligning moment and also expanded for a lateral dynamic force analysis.

The wheel dynamic model can also be utilized with a tire model to estimate the friction. In \cite{Muller2003,Hsiao2011,Rajamani2012} the wheel rolling motion is used to detect the longitudinal force and longitudinal friction adopting the powertrain configuration and wheel drive engine. The estimator is built primarily exploring the force-slip ratio plane and its relationship with the road friction coefficient.

Another model-based approach discussed in literature is a slip-slope algorithm. This method is based on the assumption that the low slip-plane zone (linear region of the force-slip plane, characterizing normal driving conditions) can be used to estimate the tire-road friction. Distinct studies \cite{Li2007,Qi2015,Xia2016} have shown this methodology.

Despite the majority of model-based methods, a number of algorithms have been studied based on different concepts to estimate the surface condition. In \cite{Casselgren2007,Tuononen2008}, an optical sensor is used as a tire sensor that can measure the road ahead and the tire carcass deflections which may be exploited in the estimation of friction potential. Cameras are also used to identify different surfaces. The detection is based on the light polarization changing when reflected from road surface \cite{Jokela2009}. Also, \cite{Jonsson2011} proposed a method that merges weather data and road images taken by a camera on the vehicle. More recently, based on the hypothesis that the friction coefficient affects the  natural frequency of the vehicle systems, such as in-wheel motor drive system or steering system, the road-friction is estimated through frequency analysis \cite{Chen2015,Chen2017}.

In the field of Machine Learning and Artificial Intelligence the work presented in \cite{Pasterkamp1998} primarily designed a feedforward neural network optimizing its topology by means of Genetic Algorithms to determine the actual tire forces from the measured signals. The estimated values were shown to have potential of exploring the nuances about TRFC.

In \cite{Matuko2008}, friction force estimation is presented in a stricted accelerating/decelerating maneuver. A two-layer feedforward neural network with linear output is used with the vehicle longitudinal dynamics. The proposed estimator is validated through a noiseless signal data.

In \cite{Zareian2015} and \cite{Zhang2017} a feedforward neural network is synthesized using the vehicle response to both longitudinal and lateral excitations. Therefore, the estimation is investigated under acceleration or brake maneuver while cornering. 

Among the model-based methods that implement TRFC estimation using neural networks, some prior works also estimate friction through road conditions. It is the case of vision-based works used to estimate the road conditions ahead of the vehicle, where the task becomes a classification problem. In \cite{Panahandeh2017}, an image dataset  was used to train three machine learning models including logistic regression, support vector machine, and neural networks to predict the friction class. In a similar form \cite{Roychowdhury2018} proposes a convolutional neural network model to learn region-specific features for road surface condition classification and how to infer the friction coefficient. Other works also implemented the classification by analyzing the road texture \cite{Yang2018,Du2019}.

In this study, the presented estimation process focuses on the dynamic characteristics of a rear-motorized-wheels electric vehicle to achieve the tire road friction estimation and contributes in the following aspect: the estimator is developed by means of a time delay neural network (TDNN) as a way to identify the TRFC based exclusively on the lateral force information and the estimates are compared with a nonlinear least squares (NLS) estimator based on a moving data window.

Although TDNN have been used for diverse purpose such as speech recognition \cite{Wu2019}, movement behavior \cite{Sarah19}, joint angle estimation \cite{Triwiyanto18} and nonlinear behavior prediction of a time series \cite{Ashwini18}, no other work has studied the TDNN as a reliable procedure to estimate or measure the effect of TRFC on a vehicle dynamic. Our main motivation for using TDNN is that the time-delay arrangement structure enables the network to effectively capture the temporal vehicle response due to a change in road friction.



This paper is organized as follows: Section \ref{sec:2} presents a vehicle theory development with mathematical models for the tire force models which are used in the estimation method. Section \ref{sec:3} details the least squares regression method. In Section \ref{sec:4}, the TDNN estimation algorithm proposed is described. Simulation results are shown and analyzed in Section \ref{sec:5}. Finally, this paper is concluded in Section \ref{sec:6}.

\section{Tire-Ground Contact Model}
\label{sec:2}

When sufficient excitations exist in the lateral direction, vehicle lateral dynamics can be used as the basis of the TRFC estimation. The most common tire friction models used in the literature are those of algebraic tire slip angle and force relationships. Although many approaches to the tire-road friction modeling can be found, for this work we selected three analytical models. These models were chosen for their clear and simple formulation. They have fewer tunning parameters and have a good representation of the tire forces nonlinearities.

\begin{figure}[htp]
	\centering 
	\includegraphics[width=1\linewidth, angle =00 ]{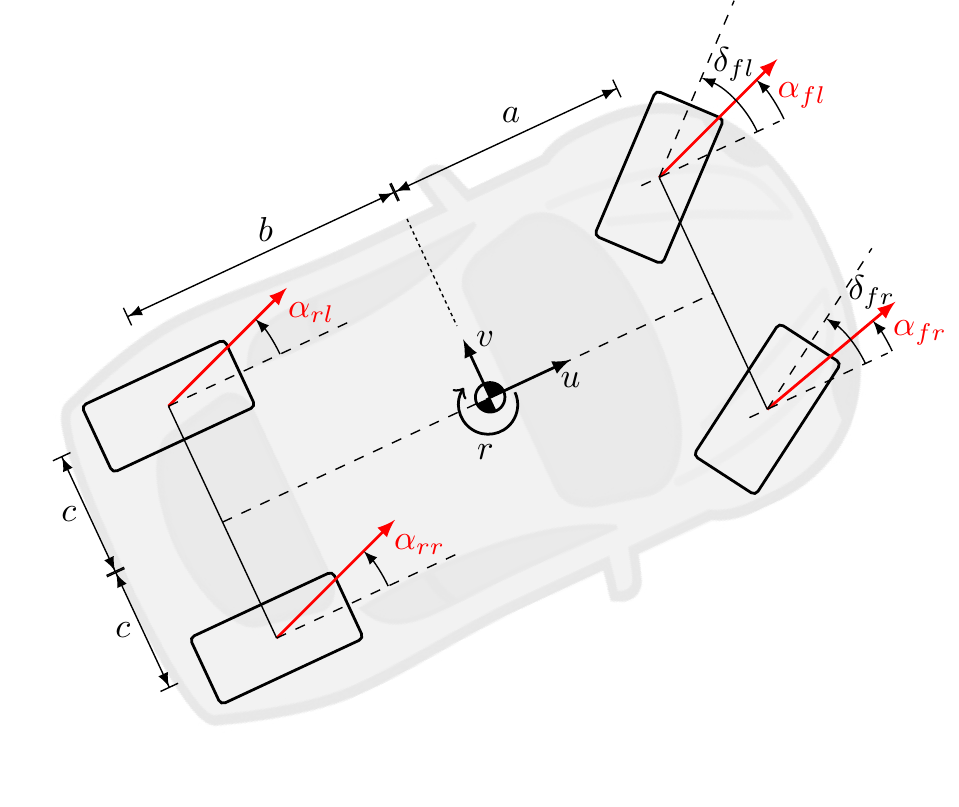}
%
	\caption{Schematic diagram of the planar vehicle and tire slip angle representation}
	\label{fig:planarvehicle} 
\end{figure}

As mentioned, the force generated between the tire and the road is related to the slip angle and it is of fundamental importance for the knowledge of how the lateral forces arise during a curve. The slip angle $\alpha$ is the angle between the orientation of the tire and the orientation of the velocity vector of the wheel, as depicted in Fig \ref{fig:planarvehicle}. 

The formal definition of slip angle can be derived via kinematic analysis of a planar four-wheel vehicle and is usually defined as:
\begin{equation}
	\begin{split}
	\alpha_{fl,fr} = & \arctan \Big( \frac{v + a r}{u \pm c r} \Big) - \delta_{fl,fr}, \\
	\alpha_{rl,rr} = & \arctan \Big( \frac{v - b r}{u \pm c r} \Big),
	\end{split}
	\label{eq:alpha}
\end{equation}
where $u$ and $v$ are vehicle longitudinal and lateral velocity components, $r$ is the vehicle yaw rate, $a$ and $b$ are the distances from the vehicle center of gravity to front and rear axles, respectively, and $c$ is half of the wheelbase distance. $\delta_i$ stand for tire steering angles and the subscripts $fl$, $fr$, $rl$ and $rr$ denote quantities corresponding respectively to the front left, front right, rear left and rear right wheels. In the vehicle model sign convention, the steering angle is negative for left turns and positive for right turns.

\subsection{Mathematical Formulation}

Tire models express the relationship between tire forces and moments with slip ratio and slip angle.

Different mathematical tire models have been developed in the literature. The most widely used model is the semi-empirical tire model introduced by Pacejka \cite{pacejka2006tyre}, called Pacejka tire model or \textit{Magic formula}. In a simplified form, the formulation of this tire model for lateral force is as follows: 
\begin{equation}
F_y = D \sin [C \arctan \{ B \alpha - E (B \alpha - \arctan B \alpha) \} ]+S_{v_y}, \\
\label{eq:pacejkalat}
\end{equation}
where D, C, B, E are the \textit{Magic formula} semi-empirical parameters based on tire measurement data, $S_{v_y}$ is the vertical offset of the characteristic curve and $\alpha$ is the slip angle.

A second model, known as Dugoff's tire model, was developed in 1969 by Dugoff et al. \cite{dugoff1969tire}. In its simplest form, the lateral force is expressed as:
\begin{align}
F_y = & -C_{\alpha} \, \tan \alpha  \,  f (\lambda), \label{eq:dugofflat}\\
\intertext{with}
f(\lambda) & = \begin{cases} (2-\lambda)~\lambda \text{,}~ &  \text{if} ~ \lambda < 1 \\ 1 & \text{otherwise}\end{cases}  \nonumber \\
\lambda & = \frac{\mu F_z}{2C_{\alpha} \lvert \tan \alpha \rvert},  \nonumber
\end{align}
where $F_z$ is the normal tire load, $\mu$ is the friction coefficient and $C_{\alpha}$ the cornering stiffness. Conceptually, cornering stiffness is a property of the tire that changes slowly with time due to tire wear, inflation pressure, and temperature fluctuations \cite{hsu2009estimation}.

Finally, another widespread model is the Brush model \cite{pacejka2006tyre}, which defines the lateral force as follows:

\begin{align}
	&F_y = \begin{cases} -3\mu F_z \theta_y \sigma_y \big\{ 1-\lvert\theta_y\sigma_y\rvert + \frac{1}{3}(\theta_y\sigma_y)^2 \big\} \text{,} &  \lvert \alpha \rvert < \alpha_{sl} \\ -\mu F_z \text{sign} (\alpha) & \text{o/w}\end{cases} 
	\label{eq:brushlat}\\
	\intertext{where}
	\normalsize
	&\theta_y = \frac{C_\alpha}{3 \mu F_z},\nonumber \\
	&\sigma_y = \tan\alpha, \nonumber\\
	&\alpha_{sl} = \arctan  \left ( \nicefrac{1}{\theta_y} \right ). \nonumber
\end{align}

Although this article only introduces the most popular and widely used approaches in tire-road friction estimation, there are many valuable studies that have tried to develop new friction models. This subject is addressed in broader texts and books about ground vehicle dynamics such as \cite{pacejka2006tyre,Wong2008theory}

\begin{figure}[h!]
	\subfloat{%
		\includegraphics[]{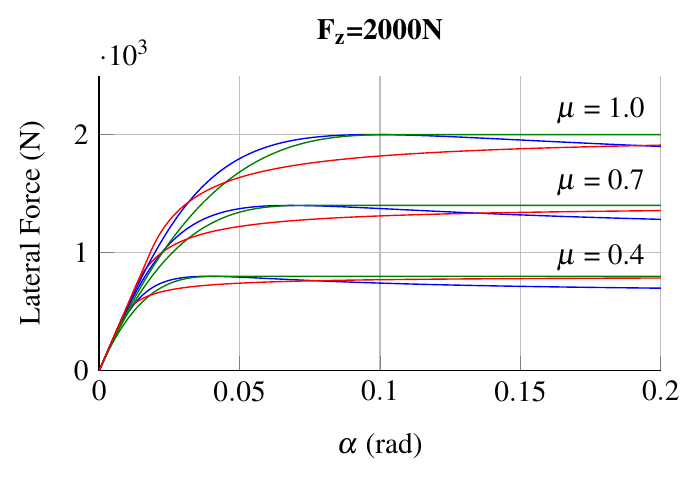}
	} \\
	\subfloat{%
		\includegraphics[]{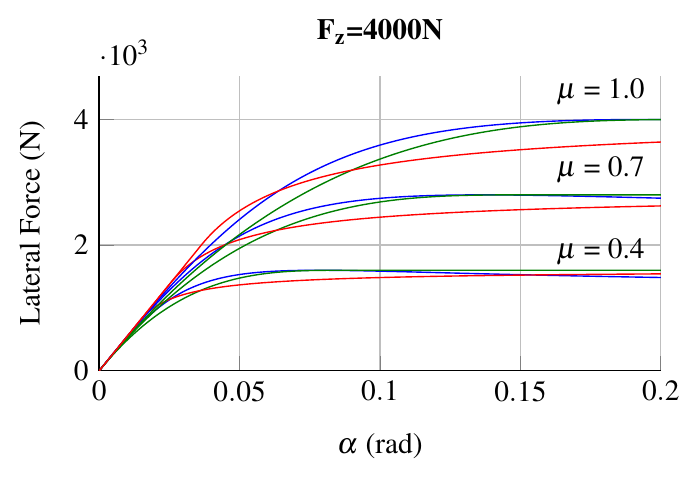}
	} \\

	\centering
	\vspace{-4mm}
	\subfloat{%
		\includegraphics[]{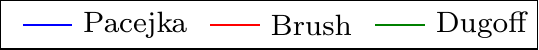}
	} \\
	\caption{Tire lateral force characteristic curve for each model, varying normal force $F_z$ and three levels of friction $\mu$}
	\label{fig:comparisonModels} 
\end{figure}

The lateral force characteristic curve for each of the presented models is shown in Fig. \ref{fig:comparisonModels} for several friction coefficients. Initially, the lateral forces increase linearly with the slip angle until it reaches saturation, which represents the tire force limits.

These models show similar behaviors when the slip angle is small. However, they may deviate from each other when high values of vertical force $F_z$ and friction $\mu$ are available. These characteristics suggest that, in the estimation processes, these models could lead to errors due to model discrepancy.

\section{Tire-road Friction Identification Through Parametric Regression}
\label{sec:3}

As seen in the previous section, the lateral force can be characterized by three fundamental parameters: tire slip angle $\alpha$, normal force $F_z$ and tire-road friction $\mu$.  

When a sufficiently large lateral excitation is detected during a vehicle maneuver, the of road friction estimation can be achieved using the measured signals and the analytical models \eqref{eq:pacejkalat}-\eqref{eq:brushlat}. This methodology can be seen as a problem of fitting experimental data to a nonlinear analytical function, as addressed in \cite{hsu2009estimation,hsuJournal,ahn2011robust,ahnJournal2013}. The method allows the formulation of the problem as one of unconstrained nonlinear least-squares (NLS) optimization. 

In other words, we desire to investigate how well we can identify our lateral tire parameters using lateral force information. This requires a good measurement of the lateral tire forces, as well as the knowledge of individual tires. However, if unavailable, the use of an estimate of the axle forces (lumped forces) may hold a lumped friction estimate.

The nonlinear curve-fitting in a least-squares problem consists of finding decision variables $x$ that solve the problem:
\begin{align}
~~~x^* &= \arg \min_x \left\lVert F(x) - \overline{F} \right\rVert _2^2, \\
&= \arg \min_x \sum_{k}^{N} \Big ( F(x) - \overline{F}_{k=1} \Big) ^2,
\label{eq:optimizationproblem}
\end{align}
where $x^*$ is the optimum value that minimizes the objective function, with $F(x)$ the parametric function and $\overline{F}$ representing the measured data. 

Assuming that the tire analytical models are a good representation of the lateral tire force behavior, they can be used as a parametric function of the NLS method with $N$ sets of observed data (in this case, groups of $F_{y}$ and $\alpha$).

Despite the promising results of this approach \cite{hsuJournal,ahnJournal2013}, the NLS method has some drawbacks. It requires a long computation time and sometimes this process fails to converge to the true optimal values. The estimator based on NLS generally shows stable estimation results, but does not always guarantee stability and it is difficult to quantify the stability and convergence \cite{ahn2011robust}. Furthermore, a critical drawback of the NLS is that it is computationally heavy. In a low-speed microprocessor, it may not sustain the same level of performance.

 As an alternative to this methodology, we propose solving the problem using neural networks. The theme is approached in a similar form, with the same window of $N$ observable data applied in a time-delayed neural network.

\section{Estimation of Tire-road Friction Coefficient Using Neural Networks}
\label{sec:4}

This section proposes a time delay neural network to detect the TRFC. Two main benefits are expected from this method: firstly, a TDNN can establish network connections and the relationship between input and output instead of storing an entire complex tire model in the controller, which can significantly reduce the computations, guarantee the real time performance and avoid model errors due to model discrepancy; secondly, because the TDNN is trained by measured data, it is able to create a mapping from input parameters to the friction coefficient and accurately capture the temporal structure hidden in the data \cite{Zhang2017}.

\begin{figure}[!thp]
	\centering 
	\includegraphics[width=1\linewidth]{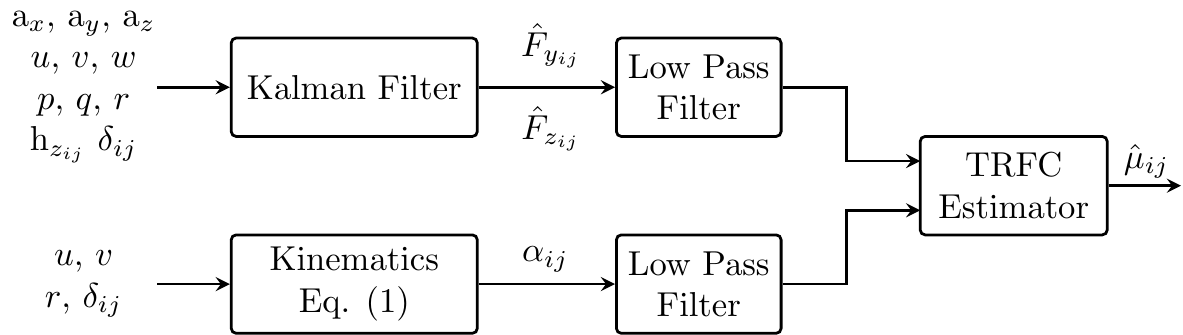}
	\caption{Block diagram of the proposed hierarchical estimator} 
	\label{fig:cascadeobs} 
\end{figure}

As the analytical models \eqref{eq:pacejkalat}-\eqref{eq:brushlat} show, the lateral force is dependent of $\alpha$, $F_z$ and $F_y$ and these are therefore the parameters selected to feed the neural network. Fig. \ref{fig:cascadeobs} shows the overall structure used for the TRFC estimation.


Although the existence of tire force sensors, the forces are still hard to measure and the sensors are very expensive. 
As solution, a Kalman filter variant is used for tire force estimation. Here, we use the approach presented in \cite{Cordeiro2016,Cordeiro2017} to estimate $F_y$ and $F_z$ using ordinary vehicle sensors, such as GPS, inertial measurement unit (IMU) and encoders. 
	
The observers are developed based on nonlinear vehicle dynamic models with the Extended Kalman Filter (EKF) algorithm. Three estimators are used in a cascade structure to decouple the forces dependency. For vertical forces a 3D vehicle model is used to create a 12-states nonlinear system. For longitudinal and lateral forces the planar vehicle model is used to create two 7-states nonlinear systems. Compromising precision and simplicity, the Dugoff tire model is chosen as the analytical model.

All measures needed for the estimation process are listed in Table \ref{tab:measures}.

\begin{table}[!ht]
	\centering
	\caption{Signals and description of measures}
	\begin{threeparttable}
		\begin{tabular}{c|c} \hline
			Signal &  Description\\ \hline \hline
			$a_x$, $a_y$, $a_z$ & \makecell{Vehicle longitudinal, lateral \\ and vertical accelerations}  \\ \hline
			\textit{u}, \textit{v}, \textit{w} & \makecell{Vehicle longitudinal, lateral \\ and vertical linear velocities} \\ \hline
			\textit{p}, \textit{q}, \textit{r} & \makecell{Vehicle roll, pitch and yaw \\ angular rates} \\ \hline
			$h_{z_{ij}}$ & Suspension deflection \\ \hline
			$\delta_{ij}$ & Tire steering angle \\ \hline \hline
		\end{tabular}
	\end{threeparttable}
	\label{tab:measures}
\end{table}

The aforementioned works about the filter and observer showed that the overall EKF forces estimator is robust against TRFC variations. It turns out that the update stage uses inertial measurements (acceleration) and is capable to correct the forces errors induced by prediction stage. The intuition behind this is that Newton's second law equations prediction is efficiently corrected by a proper acceleration measurement. Therefore, a constant $\mu = 0.8$ is applied for the EKF forces estimation.

With this approach the forces are detected individually, which holds the potential of detecting the TRFC independently for each tire. The wheel slip angle $\alpha$ is calculated straightforward using~\eqref{eq:alpha}. 

Also, a supplementary consideration should be taken to ensure the algorithm outputs reasonable estimates. Due to sensors noise and the inherent perturbation on lateral forces, and specially on slip angle calculation, the TDNN inputs should be low-pass filtered to prevent the high frequency disturbances from being propagated to the estimate. A unit gain 5 Hz low-pass filter is applied to the estimated forces and slip angle, as shown in Fig. \ref{fig:cascadeobs}. 

Before advancing into the learning process of the proposed neural network, it is important to make one addition to the model. When considering the correlation between friction and each tire measure, the correlation coefficient between the normalized lateral force $\nicefrac{F_y}{F_z}$ is significantly higher if compared with each force separately, as listed in Table \ref{tab:corrcoef}. According to \eqref{eq:alpha}, the slip angle $\alpha$ is determined by vehicle velocities and normal force $F_z$ is mainly affected by the roll over effect. These measures are only affected indirectly by friction, thereby, a low correlation is expected.
\begin{table}[!ht]
 	\centering
 	\caption{Correlation coefficients between the specified variables}
 	\label{tab:corrcoef}
 	\begin{tabular}{c|c|c|c|c}
 		\hline
 	     & $\alpha$ & $F_y$ & $F_z$ & $\nicefrac{F_y}{F_z}$\\ \hline
 		$\mu$ & 1.681e$^{-7}$ & 0.2955 & 1.186e$^{-8}$ & 0.4076 \\ \hline \hline
 	\end{tabular}
 \end{table}
 
 Given this fact, the normalized forces $\nicefrac{F_y}{F_z}$should be selected as one input to feed the neural network instead of $F_y$ and $F_z$ separately. The basis for this choice also lies on the friction circle concept in which the maximum value of the resultant force is determined along a circle (directly influenced by friction), and this value can be decomposed into the limits of the normalized forces \cite{Gim1991}.
 
 It is important to note that a positive correlation was observed for positive lateral forces (data obtained from a right hand maneuver). A negative correlation with similar magnitude is expected for negatives forces. The signal, thus, is a consequence of the reference frame.

\begin{figure}[!thp]
	\centering 
	\includegraphics[width=1\linewidth]{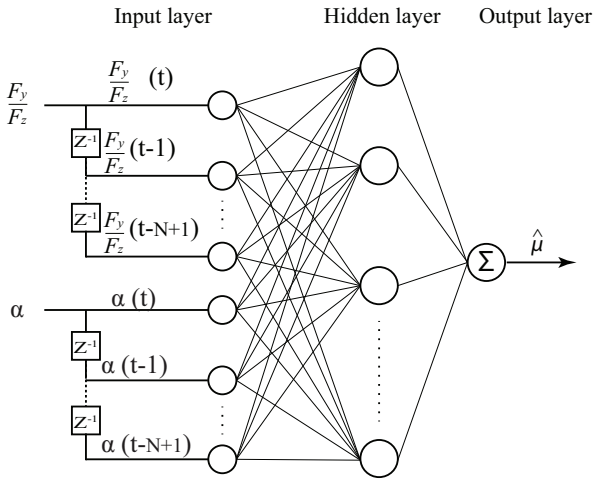}
	\caption{Time-delayed neural network architecture with a single hidden layer} 
	\label{fig:TDNNarch} 
\end{figure}

The TDNN architecture is depicted in Fig. \ref{fig:TDNNarch}. Two inputs were selected: a normalized lateral force $F_y/F_z$, obtained from the kalman estimator, and the calculated slip angle.

The configuration of the proposed TDNN for TRFC is as follows (see Fig. \ref{fig:TDNNarch}): 2 inputs with 50 samplings delay (observation window of size $\textit{N}$ = 50) and one single hidden layer with 50 neurons. The neurons differentiable transfer function is nonlinear, properly selected as a Tan-Sigmoid transfer function.


In the neural network data collecting stage, about 200,000 original data are obtained from simulation with a 100 Hz sampling rate. The range of variation of the network input parameters to the tire model is bounded as described in Table \ref{tab:rangebase}. The friction coefficient is set with different levels and the vehicle response (data of $\alpha$, $F_y$ and $F_z$) is obtained.
\begin{table}[!ht]
	\centering
	\caption{Data training parameters and space dimension}
	\begin{threeparttable}
		\begin{tabular}{c|c} \hline
			Input parameter & ~~~ Variation \\ \hline \hline
			Friction coefficient $\mu$ & ~~~~ 0.3 to 1.2 at intervals of 0.1 \\ \hline
			Slip angle $\alpha$ & ~~~~ [-0.12 0.12] \textit{rad} \\ \hline
			Lateral Force $F_y$ & ~~~~ [-2.8 2.8] \textit{kN} \\ \hline
			Normal force $F_z$ & ~~~~ [2 4.4] \textit{kN} \\ \hline \hline
		\end{tabular}
	\end{threeparttable}
	\label{tab:rangebase}
\end{table}

\begin{figure}[htp]
	\subfloat{%
		\includegraphics[width=0.49\textwidth]{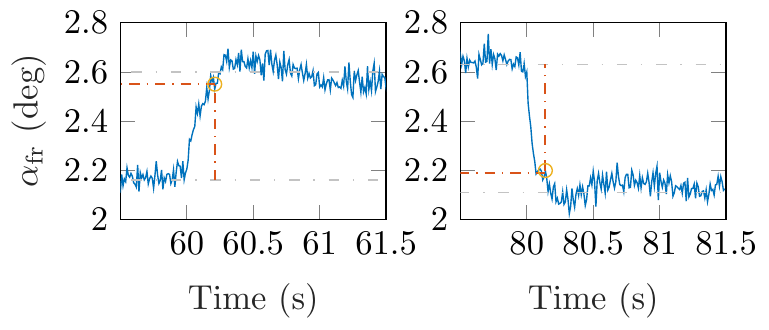}
	} 
	\caption{Rise time due to a change in friction. At instant 60 s friction changes from 0.8 to 0.6 and at instant 80 s from 0.6 to 0.9. Rise time is 0.21 s and 0.19 s for left and right plot, respectively.  }
	\label{fig:risingtime}
\end{figure}

The observation window size was chosen based on the nature of our system. The rise time response due to a change in friction is shown in Fig. \ref{fig:risingtime}. The slip angle response (and consequently lateral tire forces, see Fig. \ref{fig:comparisonModels}) shows an average rise time of 0.2 s. In an attempt to capture the vehicle behavior during these transitions, we choose the window size to be approximately double the rising time. Therefore, it is defined N = 50 that corresponds to an observation window of 0.5 seconds.

As in the classical Neural Network, the Time Delay Neural Network also has a training phase. Training was achieved using Matlab Neural Network Toolbox. The \textit{Levenberg-Marquardt} algorithm is used with 1000 epochs of training iterations with 70\% of the collected data randomly taken as the training set, 15\% used for validation set and 15\% as the test set.

Training phase and further evaluations were carried out on a desktop architecture which features a four-core 4.00 GHz Intel Core i7-6700K Processor, NVIDIA GeForce GTX 950, 32GB of RAM and Ubuntu 16 LTS OS. The system took 7002 seconds to train the aforementioned dataset.
	
\begin{figure}[htp]
	\centering 
	\includegraphics[width=1\linewidth]{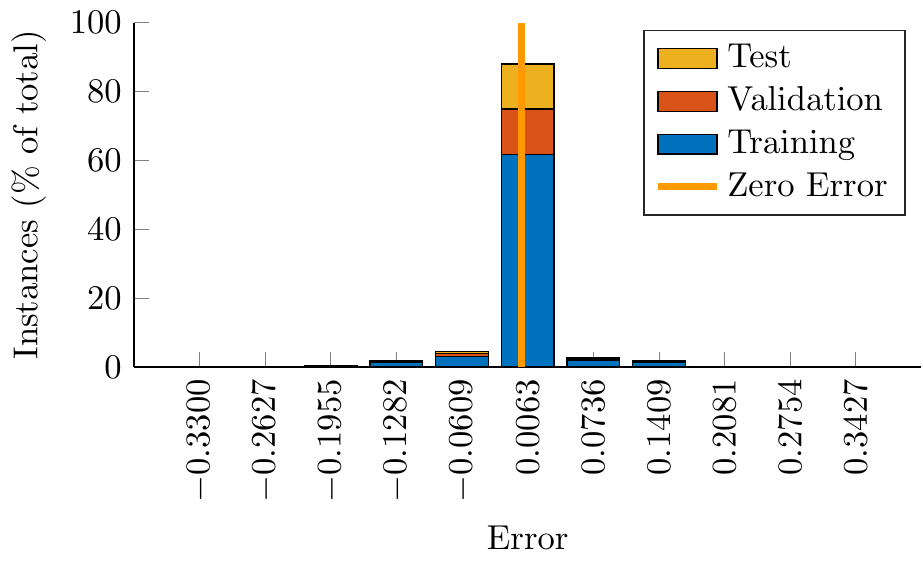}
	\label{fig:roadlayouta} 	
	\caption{Error histogram of the TDNN dataset}
	\label{fig:histogramerror} 
\end{figure}

Fig. \ref{fig:histogramerror} shows the error residuals of the trained neural network. About 88\% of the total data are distributed around zero.

The correlation coefficient (R-value), which is a linear regression between the TDNN predicted values and the targets, and the mean squared error (mse) are shown in Table \ref{tab:RValue}. With R values around 0.94 and 0.95, suggests that the models has a considerable potential of matching the network response to the friction change patterns. Notice also that both R and mse values are close for Training, Validation and Test dataset, which indicates that the network is not overfitting the data. 

\begin{table}[!ht]
	\centering
	\caption{Correlation coefficient R-value and mse error. Values of R closer to 1 indicate better agreement between targets and predicted values}
	\label{tab:RValue}
	\begin{tabular}{ccc}
		\hline
		& R & mse \\ \cline{2-3} 
		Training & 0.94355  & 1.0680e$^{-3}$    \\ \cline{2-3} 
		Validation & 0.94123  & 1.2597e$^{-3}$    \\ \cline{2-3} 
		Test & 0.94059  &  1.3500e$^{-3}$   \\ \cline{2-3} 
		Total & 0.94276  &  1.1391e$^{-3}$   \\ \hline \hline
	\end{tabular}
\end{table}

\section{Results}
\label{sec:5}

The simulation results presented in this section are obtained using Matlab/Simulink. A representative and realistic full-vehicle multibody dynamics model (including a steering system, powertrain system, suspension system and the Pacejka tire model for tire ground interactions), was used consisting of the following motions:
\begin{itemize}
	\item  Longitudinal, lateral and vertical body motion;
	\item Wheels rotation;
	\item Unsprung masses motion;
	\item Pitch, roll and yaw body rotation.
\end{itemize}

The physical parameters of the car used in this study are listed in Table \ref{tab:physicalparam}. All values are extracted from \cite{rafaelcordeirothesis} where a complete vehicle modeling and data validation are given. The simulator dynamics is formulated as a 32-states model.
The vehicle is configured to simulate a independent rear wheel drive vehicle, providing references of a vehicle state and measured signals. Gaussian noises are added (according to the commercial MTi Xsens sensor specifications (MTi-G-700)) in the simulated measurements to realistically reproduce a real application.

\begin{table}[!ht]
	\centering
	\caption{List of vehicle main physical parameters \cite{rafaelcordeirothesis}}
	\begin{threeparttable}
		\begin{tabular}{c|c} \hline
			Parameter name & ~~~ Value \\ \hline \hline
			Vehicle mass & ~~~~ 1100 kg \\ \hline
			Yaw inertia moment & ~~~~ 1350 \nicefrac{kg}{$m^2$} \\ \hline
			Roll inertia moment & ~~~~ 337.5 \nicefrac{kg}{$m^2$} \\ \hline
			Pitch inertia moment & ~~~~ 1350 \nicefrac{kg}{$m^2$} \\ \hline
			Distance from CG to front wheels & ~~~~ 1.5 m \\ \hline 
			Distance from CG to rear wheels & ~~~~ 1.9 m  \\ \hline
			Wheelbase & ~~~~  1.8 m\\ \hline
			Wheel rotational inertia & ~~~~ 1 \nicefrac{kg}{$m^2$} \\ \hline
			Wheel radius & ~~~~ 0.25 m  \\ \hline
			Height of CG & ~~~~ 0.5 m  \\ \hline \hline
		\end{tabular}
	\end{threeparttable}
	\label{tab:physicalparam}
\end{table}

\begin{figure}[htp]
	\centering 
	\subfloat[]{%
		\includegraphics[width=.3\linewidth]{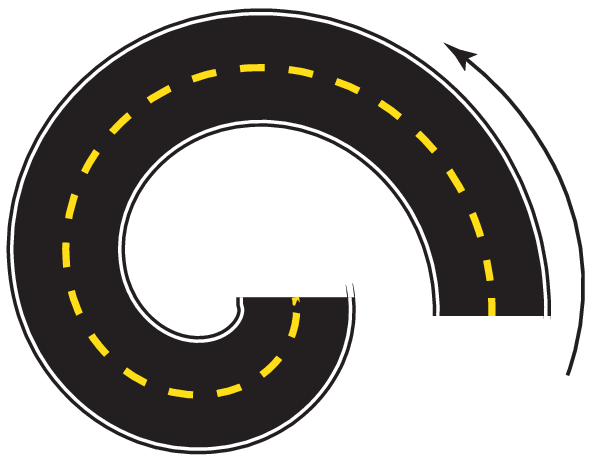}
		\label{fig:roadlayouta} 
	}	
	\subfloat[]{%
		\includegraphics[width=.3\linewidth]{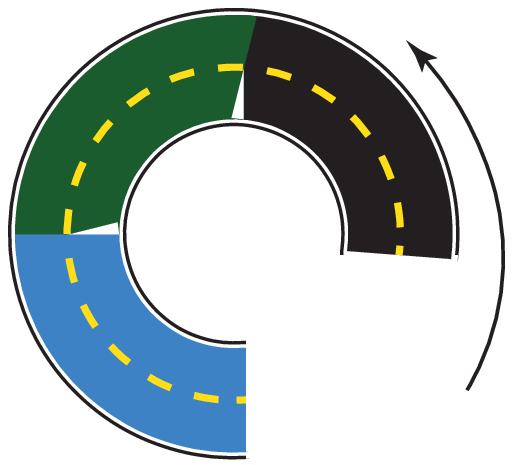}
		\label{fig:roadlayoutb} 
	}
	\subfloat[]{%
		\includegraphics[width=.3\linewidth]{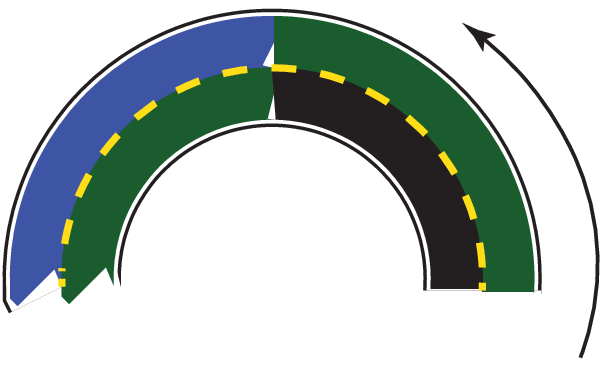}
		\label{fig:roadlayoutc} 
	}
	\caption{Schematic road layout with time-dependent changing of friction coefficient }
	\label{fig:roadlayout} 
\end{figure}

The simulation results of three representative maneuvers are presented here. Table \ref{tab:purposemaneuver} gives the details and purpose of each maneuver. Fig. \ref{fig:roadlayout} provides the physical representation of each proposed scenario where the color designates the friction coefficient. The change in friction is time-dependent and occurs at successive time intervals.
	
During simulation, the system is set on an equilibrium point with constant longitudinal velocity. It is achieved by a PID cruise controller \cite{Osman09}. In fact, with constant longitudinal speed, the longitudinal tire forces are reactive with low magnitude and would not have significant impact on lateral forces due coupling effect \cite{Hindiyeh09}.

Each case was performed on a different theoretical surface, where a theoretical $\mu$ = 1.0 surface roughly corresponds to driving on a dry pavement, $\mu$ = 0.8 on a wet pavement and $\mu$ = 0.6 corresponds to driving on gravel \cite{Wang2004}.

\begin{table*}[]
	\centering
	\caption{Details and investigative purpose of each simulated maneuver}
	\label{tab:purposemaneuver}
	\begin{tabular}{c|c|c}
		\hline
		Simulated Maneuver                                    & Test Surface & Purpose \\ \hline \hline
		 \makecell{Ramp steer with constant \\ friction coefficient \\ (Fig. \ref{fig:roadlayouta}) } & \makecell{Constant dry pavement \\ ($\mu$ = 0.8)}    &  \makecell{Investigate the early friction sensing ability of the \\ estimator. During the maneuver, lateral forces are \\ gradually increasing, eventually reaching saturation.} \\ \hline
		\makecell{Constant steer - case 1: \\ step varying $\mu$ for all tires \\ (Fig. \ref{fig:roadlayoutb})}     & \makecell{Five levels of TRFC \\ with unequal steep sizes \\  ($\mu$ = 1.0 to 0.6)}   &  \makecell{Determine the estimator response to a quick \\ change in road surface. Friction coefficient is \\ varied at five new levels.} \\ \hline
		\makecell{Constant steer - case 2: \\ different step varying $\mu$ \\ for left/right tires  \\ (Fig. \ref{fig:roadlayoutc})} & \makecell{Each wheel is subjected \\ simultaneously to two \\ different frictions \\ ($\mu_{FL}$ = 0.9, 0.8, $\mu_{FR}$ = 0.8, 0.7)}     &  \makecell{Validate the TRFC detection under small friction \\variations. Also, each wheel is exposed to different \\ surfaces at the same time (distinct $\mu$ for \\ FL and FR tires).}\\ \hline \hline
	\end{tabular}
\end{table*}

It is valid to remark that, even though our training and test sets included both left and right turns, due to the symmetry of the system, right turns have a mirror response of left turns that is also included in the network as the result of a data augmentation process.

\subsection{Ramp Steer Maneuver}

\begin{figure}[htp]
	\subfloat{%
		\includegraphics[]{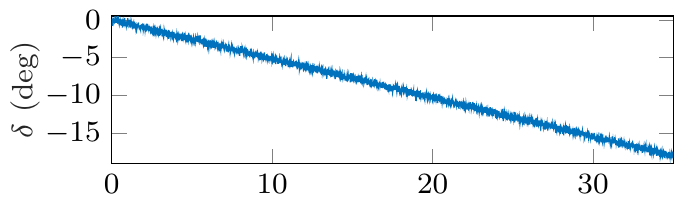}
	} \\
	\subfloat{%
		\includegraphics[]{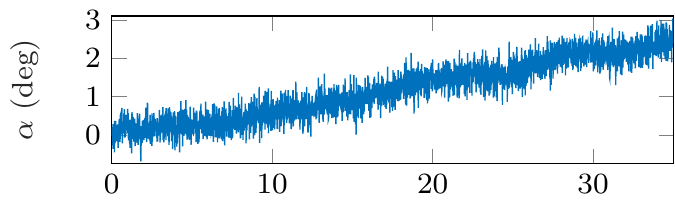}
	} \\
	\subfloat{%
		\includegraphics[]{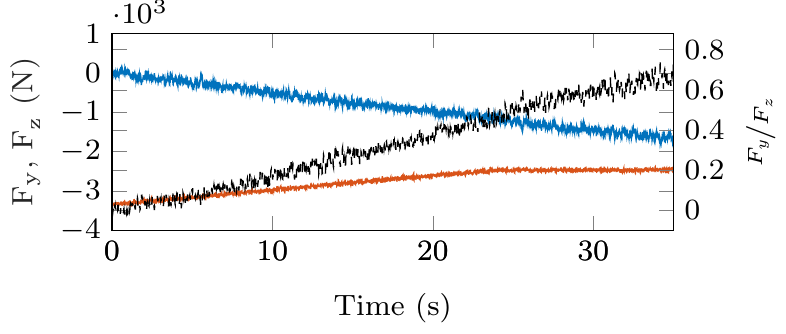}
	} 
	
	\centering
	\vspace{-3mm}
	\subfloat{%
		\includegraphics[]{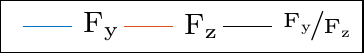}
	} \\
	\caption{Front right lateral force and slip angle for a ramp steer maneuver}
	\label{fig:rampsteermaneuver} 
\end{figure}

Fig. \ref{fig:rampsteermaneuver}  displays the data resulting from the simulation of a left-hand ramp steer maneuver. The steering angle $\delta$ goes linearly from 0 to -18 degrees at the roadwheels reference. 

To ensure that there is enough data to be meaningful for the NLS fit and the TDNN approach, first the process is initialized by placing a slip angle threshold $\alpha_{thres}$. The slip angle data must exceed $\alpha_{thres}$ before the estimation begins, otherwise the fitting optimization may not guarantee a reliable solution.

The TDNN estimator will be here compared with the NLS approach. To show the dependency of the NLS fit with the mathematical model, the regression is performed choosing the Dugoff and Brush models as a parametric function of the nonlinear regression. The window is selected with size $N=50$ and will be used in all cases showed from here on.

\begin{figure}[!ht]
	\centering
    \includegraphics[]{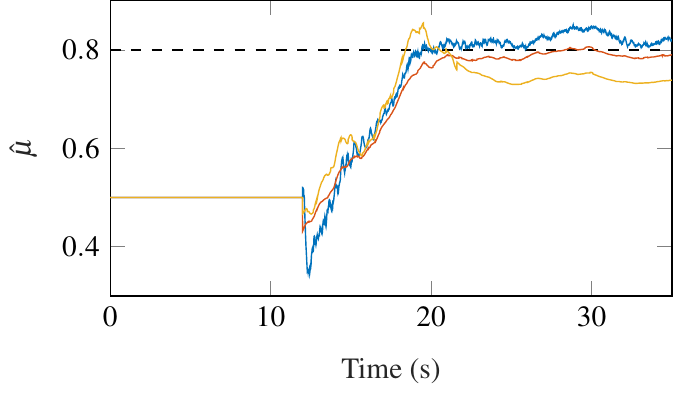}
    \subfloat{%
    	\includegraphics[]{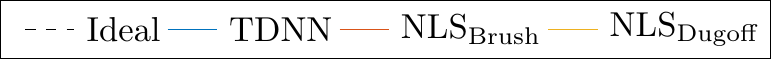}
    } \\
	\caption{Friction estimates from ($F_y$,$\alpha$) data for different approaches}
	\label{fig:rampmaneuverestimates} 
\end{figure}

Using the estimation algorithm with $\alpha_{thres}$ = 1 degree, the NLS and TDNN algorithm waits until the front tire slip angle exceeds $\alpha_{thres}$ at t = 12 s before fitting the force-slip data (see Fig. \ref{fig:rampmaneuverestimates}). 
The estimated value $\hat{\mu}$ is the optimum solution $x^*$ of the optimization problem \eqref{eq:optimizationproblem} and before instant 12 s the estimation simply holds $\hat{\mu}$ at initial value $\mu_0 = 0.5$.

The slight increase in the friction estimate as the maneuver progresses is expected. Initially, during linear tire regime operation, lateral forces measurements have yet to reach their peak value and both methods underestimate the friction coefficient. As more lateral force measurements become available, the peak force limit is reached and the friction estimate reaches a final estimate. Therefore, adequately large slip angles are required for stable and accurate estimation in both TDNN and NLS methods and the slip angle data threshold is indispensable.

The model error also becomes apparent on the NLS fit, where Brush and Dugoff models show different convergence values due to model discrepancy. This divergence arises due to the difference between Pacejka model, used to generate the data, with the models used on the estimators (as discussed in section \ref{sec:2}).

\subsection{Constant Steer: Case 1}

\begin{figure}[htp]
	\subfloat{%
		\includegraphics[]{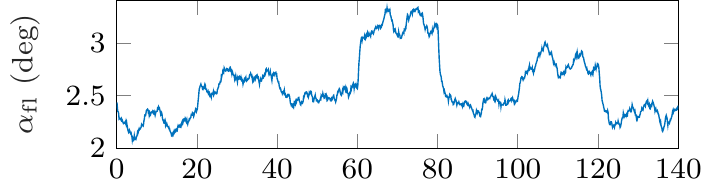}
	} \\
	\subfloat{%
		\includegraphics[]{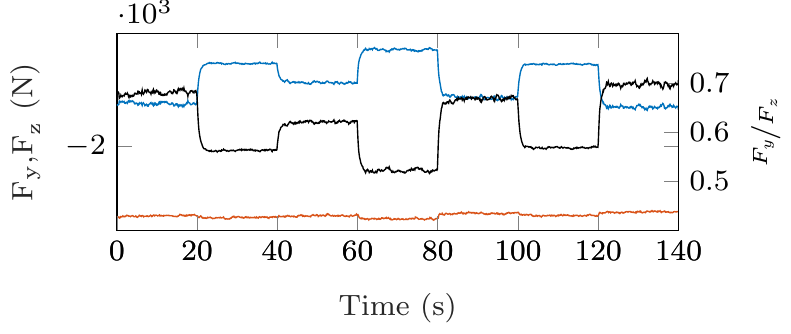}
	}
	\begin{center}
		\vspace{-6mm}
		\setcounter{subfigure}{0}%
		\subfloat[front left (fl)]{%
			\includegraphics[]{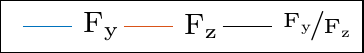}
		} \\
	\end{center}
	\subfloat{%
		\includegraphics[]{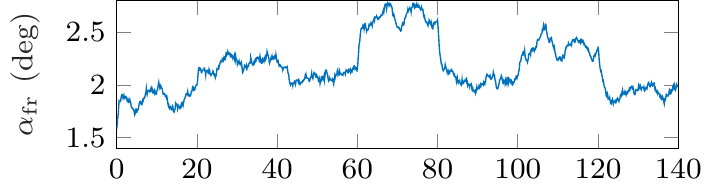}
	} \\
	\subfloat{%
		\includegraphics[]{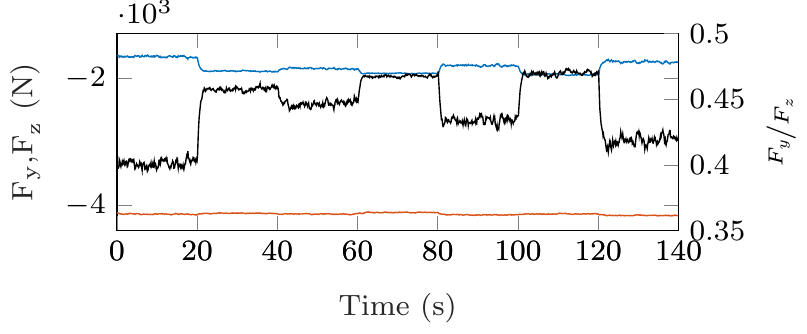}
	} \\
	\begin{center}
		\vspace{-9mm}
		\setcounter{subfigure}{1}%
		\subfloat[front right (fr)]{%
			\includegraphics[]{Fig10c.pdf}
		} \\
	\end{center}
	\caption{Slip angle, vertical and lateral forces of the tires: Constant steer case 1}
	\label{fig:latexcitation}
\end{figure}

In this setup, the tire-road friction coefficient is set at five levels, varying randomly from 1.0 to 0.6. The transitions occur during successive equal time intervals of 10 seconds. The vehicle is set on an equilibrium point in a constant left turn maneuver. The steering angle of the front left and right tires is set to -18.36 and -15.82 degrees. These values follow the steering Ackerman Geometry. As a consequence, the inner-turn wheel reaches higher tire side slip and lateral force if compared to the outer-turn wheel, as shown in Fig. \ref{fig:latexcitation}. Also, vertical force is higher on the right tire due to the load transfer. This response appears as a result of the roll over effect measured as a change in vehicle center of mass location relative to the wheels.

\begin{figure}[!htp]
	\subfloat{%
		\includegraphics[]{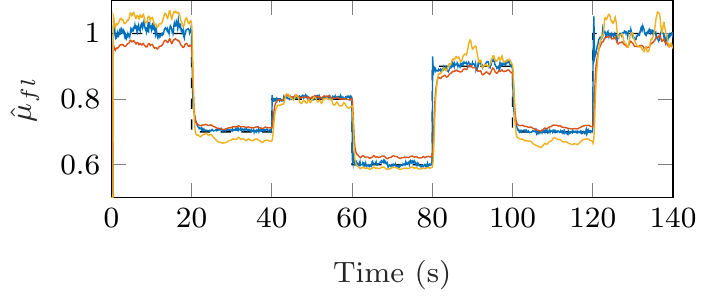}
		\label{fig:timevaryingfriction_fig9a} 
	} \\
	\subfloat{%
		\includegraphics[]{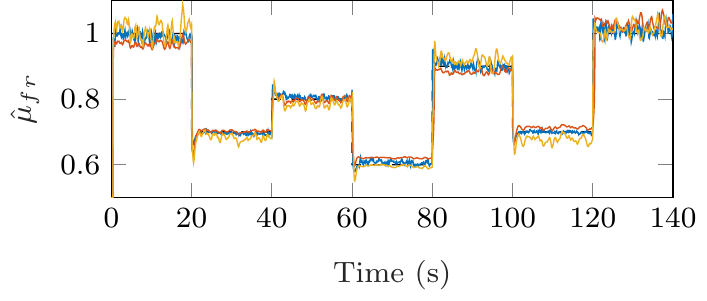}
	} \\
	\centering
	\vspace{-7mm}
	\subfloat{%
		\includegraphics[]{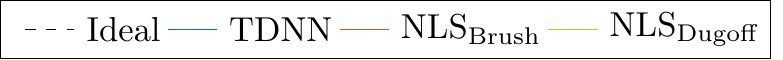}
	} \\
	\vspace{7mm}
	\caption{Front left and front right friction estimates in a time varying friction scenario}
	\label{fig:timevaryingfriction} 
\end{figure}

The TRFC estimation results are shown in Fig. \ref{fig:timevaryingfriction}. One can note the front left estimative is more accurate and less oscillating than the front right estimative. A necessary condition for good estimation results, as shown previously, is a large lateral excitation (high slip angle). 

This scenario highlights the dependency of the NLS to the parametric function. The NLS final estimates of each interval show a constant error bias while the TDNN produces a solid and concise estimative. Naturally, there are discrepancies between the dynamic behavior of the real tire system and the derived mathematical model (see Fig. \ref{fig:comparisonModels}). Therefore, a constant error bias should be expected on the NLS model-based approach.

Table \ref{tab:case1} displays the root mean square (RMS) error of the estimates of the front left and right tires. Although very similar, a high overall estimation accuracy is achieved for both techniques and it shows great promise for a real implementation.

\begin{table}[htbp]
	\centering
	\caption{RMS error of the estimated friction - case 1}
	\begin{tabular}{c|c|c}
		\hline
		Methodology    &  $\mu_{fl}$ & $\mu_{fr}$ \\ \hline \hline
		TDNN            & 0.0346 & 0.0350 \\ \hline
		$\text{NLS}_{\text{Dugoff}}$ & 0.0421 & 0.0357\\ \hline
		$\text{NLS}_{\text{Brush}}$  & 0.0546 & 0.0395\\ \hline \hline
	\end{tabular}
	\label{tab:case1}
\end{table}

\begin{figure}[htp]
	\subfloat{%
		\includegraphics[width=0.49\textwidth]{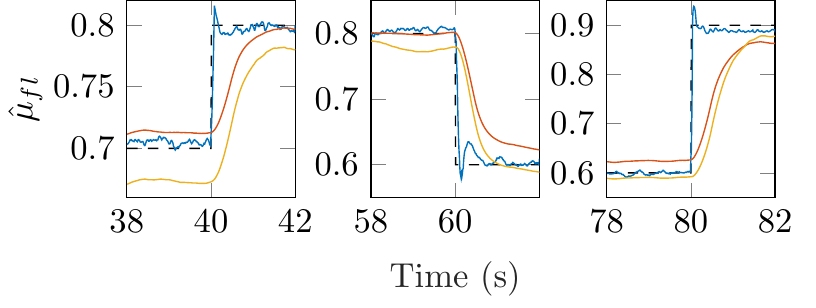}
	} \\
	\begin{center}
		\vspace{-7mm} 
		\hspace{2mm}
		\setcounter{subfigure}{0}%
			\includegraphics[]{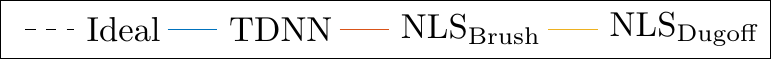}
	\end{center}
	\caption{Friction estimates rate of convergence comparison}
	\label{fig:rateconvergence}
\end{figure}

Moreover, the rate of convergence is slightly higher on the TDNN approach. Fig. \ref{fig:rateconvergence} highlights this response by zooming in Fig. \ref{fig:timevaryingfriction_fig9a} on three intervals of friction transitions. The TDNN estimates converges to a more accurate values with faster responses than the NLS estimates. This behavior lies in the fact that the relationship between input and output was correctly mapped on the database and therefore can be observed on the following results.

\subsection{Constant Steer: Case 2}

While case 1 showed an equal change for both wheels, here, the tire-road friction coefficient is set at two different levels for each tire. The transition occur during an equal time interval of 10 seconds. Left wheel friction undergoes a transition from 0.9 to 0.8 while right TRFC goes from 0.8 to 0.7. 

The vehicle is set on an equilibrium point in a constant left turn maneuver. The knowledge of forces and friction of individual tires is desirable and would offer stability control systems with most needed information. Thus, this change maneuver is conducted to verify that the estimator can identify the friction for each tire individually and certify that the estimation is indeed independent for each wheel.

\begin{figure}[htp]
	\subfloat{%
		\includegraphics[]{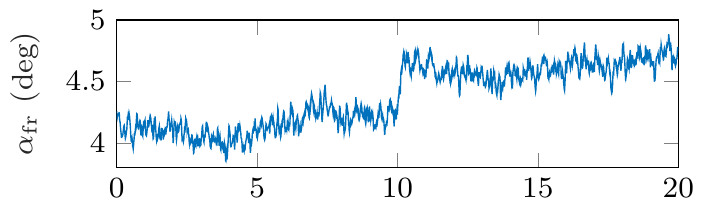}
	} \\
	\subfloat{%
		\includegraphics[]{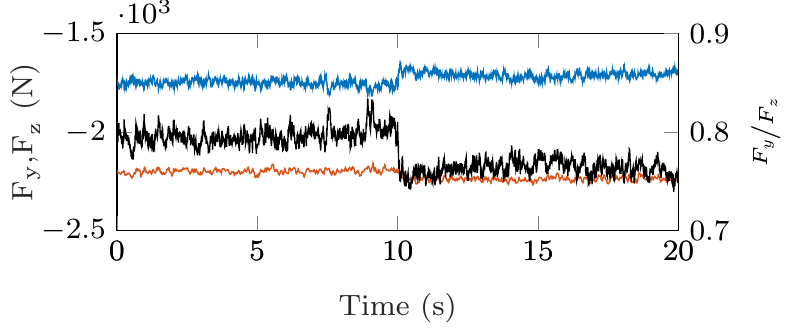}
	}\\
	\begin{center}
		\vspace{-7mm}
		\setcounter{subfigure}{0}%
		\subfloat[front left (fl)]{%
			\includegraphics[]{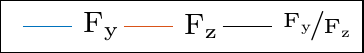}
		} \\
	\end{center}
	\subfloat{%
		\includegraphics[]{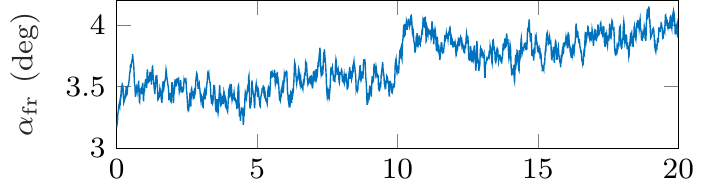}
	} \\
	\subfloat{%
		\includegraphics[]{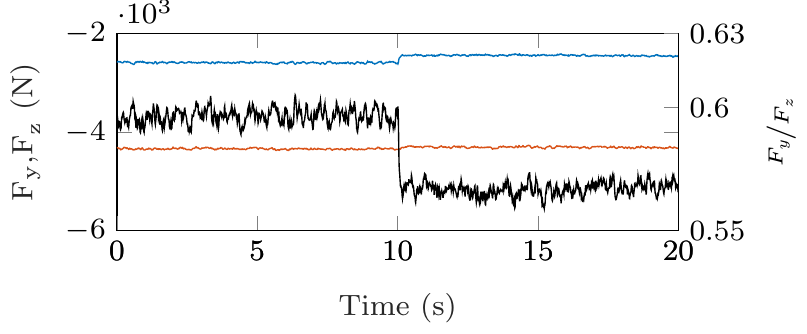}
	}
	\begin{center}
		\vspace{-7mm}
		\setcounter{subfigure}{1}%
		\subfloat[front right (fr)]{%
			\includegraphics[]{Fig13c.pdf}
		} \\
	\end{center}
	\caption{Slip angle, vertical and lateral forces of the tires: Constant steer case 2}
	\label{fig:distlatexcitation2}
\end{figure}

Fig. \ref{fig:distlatexcitation2} displays the simulated slip angle, lateral and vertical forces from the proposed right-hand steer maneuver. At instant 10 s, the road surface adhesion coefficient decreases to a different value for each tire. Again, vertical forces are maintained constant due to the maneuver nature, with the changing in friction mostly affecting lateral forces and slip angle.

\begin{figure}[!htp]
	\subfloat{%
		\includegraphics[]{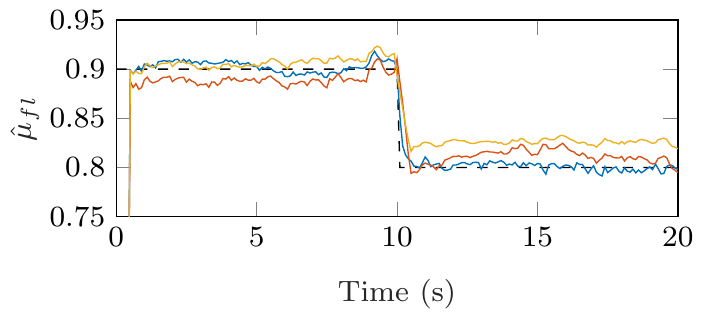}
	} \\
	\subfloat{%
		\includegraphics[]{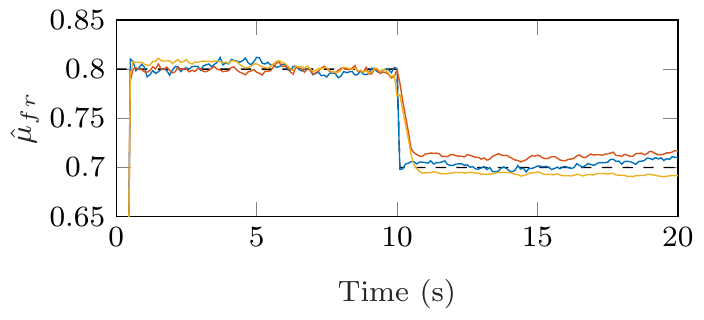}
	} \\	
	\centering
	\subfloat{%
		\includegraphics[]{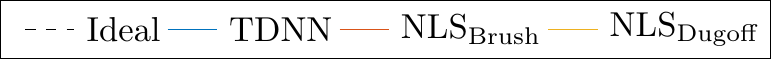}
	} 
	\vspace{7mm}
	\caption{Front left and front right friction estimates in a time varying friction scenario with different friction for each wheel}
	\label{fig:distinctfriction} 
\end{figure}

Fig. \ref{fig:distinctfriction} shows the estimation results. With the TDNN approach, the individual wheel friction is confidently estimated with high accuracy. Since slip angles are larger than for the case 1 scenario, it satisfies the required large lateral forces excitation and gives very accurate estimates. On the NLS estimation results, however, a constant error are still apparent and can be seen during the transitions. This characteristic should be considered when the estimated result is used for control purposes.

The TDNN shows a slightly better estimation quality, as seen in the RMS error listed in Table \ref{tab:case2}. Note that the regression based method still exhibits the hindsight bias caused by the inevitable model differences.

\begin{table}[htbp]
	\centering
	\caption{RMS error of the estimated friction - case 2}
	\begin{tabular}{c|c|c}
		\hline
		Methodology    &  $\mu_{fl}$ & $\mu_{fr}$ \\ \hline \hline
		TDNN            & 0.0638 & 0.0477 \\ \hline
		$\text{NLS}_{\text{Dugoff}}$ & 0.0652 & 0.0493\\ \hline
		$\text{NLS}_{\text{Brush}}$  & 0.0666 & 0.0486\\ \hline \hline
	\end{tabular}
	\label{tab:case2}
\end{table}

\begin{figure}[htp]
	\centering
	\subfloat{%
		\includegraphics{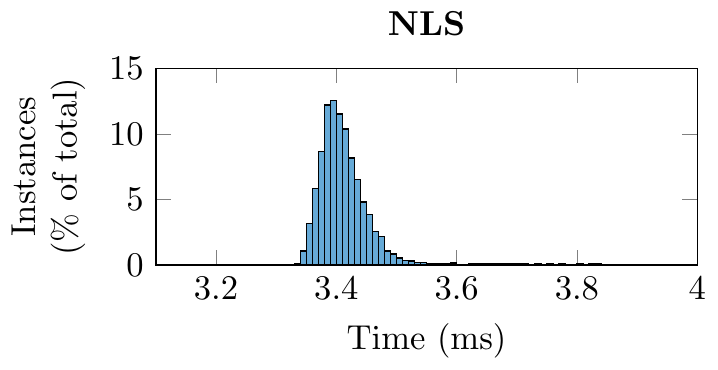}
	} \\
	\vspace{-3.5mm}
	\subfloat{%
		\includegraphics{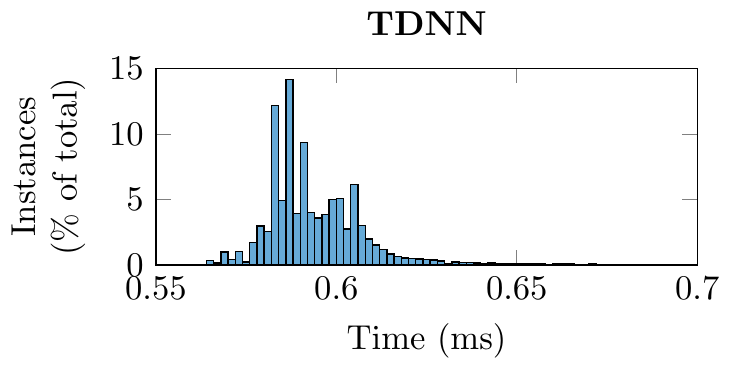}
	} \\
	\caption{Normalized histogram of computational time for NLS and TDNN methods}
	\label{fig:histogramtime} 
\end{figure}

A significant difference in processing time is found when comparing performance and computational efficiency. Fig. \ref{fig:histogramtime} shows the normalized histogram of the computation time for both techniques. These values, computed at each time instant, were obtained from 14000 data signals.

It is clear that the computation time of TDNN algorithm is lower than NLS. In summary, TDNN averages 0.594 milliseconds to process the signal (corresponding to one iteration) compared to 3.379 milliseconds of the NLS. This has the potential to allow a higher sample frequency of the complete system.
	
\section{Conclusions}
\label{sec:6}


In this study, we presented a hierarchical TRFC estimation method based on a time delay neural network and compared it with a classical nonlinear regression approach, using the same data observation window. The overall estimation algorithm was evaluated on varying road surfaces with three different scenarios. A high-fidelity and realistic full-vehicle multibody dynamics model for Matlab/Simulink platform was used in the simulations.

Although road-friction was accurately identified using both algorithms, there is a primary shortcoming in the presented lateral-force based friction estimation: it requires sufficient levels of lateral excitation for the correct friction identification. An earlier knowledge of the TRFC is desirable, however, both approaches showed a similar behavior: a necessary waiting time for the tire slip angle to fulfill the observation window satisfying the specified excitation threshold. 

Nonetheless, as an algebraic methods, the NLS method relies more heavily on an accurate model which may be a source for estimation errors. Indeed, the NLS needs a prespecified parametric function for precise tunning parameters and, as a consequence, stationary estimation errors may be expected if the given function is not adequate.

On the other hand, the TDNN method is independent of any mathematical tire model, however, requires a sufficient and representative database. In this study, the TDNN was also able to provide estimates with lower RMS errors compared with the classical regression approach. It also demands less computation time at each time instant and may be the best alternative for a real time implementation in embedded systems.

Since the proposed method is only analyzed theoretically and validated via simulation, an actual benchmark or field test is needed in the subsequent work to verify the proposed approach. Future works may also include the design of a time delay neural network containing not only lateral information but also longitudinal forces, slip ratio and self-moment align.
%

\begin{acknowledgements}
This work was supported by FCT Portugal, through IDMEC under projects LAETA (UID/EMS/ 50022/2019). The mobility of A. Ribeiro has been possible with the Erasmus Mundus SMART$^2$ support (Project Reference: 552042-EM-1-2014-1-FR-ERA MUNDUS-EMA2) coordinated by CENTRALESUPELEC. The authors also acknowledge the support of FAPESP through Regular project AutoVERDE N. 2018/04905-1, Ph.D. FAPESP 2018/05712-2 and CNPq grant 305600/2017-6.
\end{acknowledgements}

\bibliographystyle{unsrt}      
\bibliography{references}

\end{document}